\documentclass[letterpaper]{article} 
\usepackage{aaai2027}  
\usepackage[hyphens]{url}  
\usepackage{graphicx} 
\usepackage{natbib}  
\usepackage{caption} 
\usepackage{algorithm}
\usepackage{bbding}
\usepackage{color}
\usepackage{graphicx}
\usepackage{xcolor}
\usepackage{colortbl, booktabs}
\usepackage{multicol}
\usepackage{multirow}
\usepackage{comment}
\usepackage{bm}
\usepackage{amssymb}
\usepackage{amsmath}
\usepackage{dsfont}
\usepackage{subcaption}
\usepackage{booktabs}
\usepackage{graphicx}
\usepackage{threeparttable}
\usepackage{algpseudocode}
\usepackage{float}
\usepackage{newfloat}
\usepackage{listings}
\DeclareCaptionStyle{ruled}{labelfont=normalfont,labelsep=colon,strut=off} 
\floatstyle{ruled}
\newfloat{listing}{tb}{lst}{}
\floatname{listing}{Listing}

\usepackage{booktabs}

\title{Recursive Vision Language Models for General Symbolic Reasoning}
\author{
    Omid Nejati Manzari\textsuperscript{\rm 1},
    Guillaume Lajoie\textsuperscript{\rm 2,\rm 3},
    Hassan Rivaz\textsuperscript{\rm 1}\corresponding
}
\affiliations{
    \textsuperscript{\rm 1}Concordia University, Montreal, Canada
    \\

    \textsuperscript{\rm 2}Mila — Quebec AI Institute, 
     \textsuperscript{\rm 3}Université de Montréal
    
}

\begin{document}

\maketitle
\begin{abstract}
Hard symbolic-reasoning tasks such as Sudoku, maze pathfinding, and ARC remain challenging for LLMs due to their fixed-depth autoregressive reasoning, which limits systematic search, refinement, and backtracking. While recursive models such as Hierarchical Reasoning Model (HRM) and Tiny Recursive Model (TRM) address this limitation through iterative latent-state refinement, they are typically task-specific and do not leverage pretrained language priors.
We propose R-Qwen, a recursive reasoning framework built upon a pretrained Qwen backbone. 
R-Qwen repeatedly refines a candidate solution through programmatic self-recursion and deep supervision, combining the structured iterative computation of recursive models with the linguistic and reasoning priors of pretrained LLMs. We further adapt Hierarchical Supervision Weighting (HSW) to autoregressive models by exponentially weighting losses across recursive steps. HSW reduces gradient variance by at least 50\%, improves the signal-to-noise ratio of stochastic gradients, and accelerates convergence. Across eight challenging benchmarks, R-Qwen consistently outperforms prior recursive reasoning models and substantially larger LLMs while using a comparable number of trainable parameters. Notably, on ARC-AGI dataset, our model achieves a 27.6\% improvement over the baseline, highlighting the effectiveness of recursive refinement for general symbolic reasoning. These results suggest that recursive reasoning mechanisms and pretrained language model priors are complementary approaches for improving symbolic puzzle-solving. Code and models will be released after acceptance.
\end{abstract}

\begin{links}
    \link{Code}{https://github.com/IMPACT-L/RQwen}
\end{links}

\section{Introduction}
\label{sec:intro}

Large language models (LLMs) have advanced rapidly in knowledge- and
language-intensive tasks, yet they remain unreliable on symbolic puzzles that
require systematic search, backtracking, and exact structured outputs, including
Sudoku, Maze pathfinding, and the Abstraction and Reasoning Corpus
(ARC-AGI)~\citep{chollet2019arc,arcprize2025arcagi2}. Autoregressive generation
can propagate a single early error through an entire solution, while the fixed
computational depth of standard Transformers limits the iterative procedures
they can execute end-to-end~\citep{jolicoeurmartineau2025trm,wang2025hrm}.
Chain-of-thought prompting and additional test-time computation help on many
reasoning tasks, but remain brittle on search-heavy problems and often require
substantial reasoning supervision~\citep{wang2025hrm,jolicoeurmartineau2025trm}.

Recursive reasoning offers a complementary approach by repeatedly refining an
intermediate solution. Models such as HRM and TRM demonstrate that small
recursive networks can solve difficult symbolic tasks by increasing
computational depth rather than parameter count
\citep{wang2025hrm,jolicoeurmartineau2025trm}. Related approaches introduce
looped computation, fixed-point refinement, or recursive self-calling
\citep{jeddi2026loopformer,fan2026bridging,feinashley2026solve,zhang2026rlm}.
However, these methods generally operate in latent space, require specialized
architectures, target a narrow task family, or recurse over prompts rather than
explicit candidate solutions. This leaves open whether a pretrained
autoregressive model can learn a reusable recursive refinement policy without
modifying its backbone.

We introduce \textbf{R-Qwen}, a task-general recursive search-and-refine
framework built on a pretrained Qwen backbone. At each step, the model receives
the original problem and its current serialized candidate, generates an improved
candidate, and applies a task-specific projection that preserves hard constraints
before the next recursion. The same refinement operator is used across Sudoku,
Maze, ARC, and crosswords. R-Qwen is trained parameter-efficiently with LoRA
and deep supervision, together with Hierarchical Supervision Weighting
(HSW)~\citep{qasim2025cgar}.

Recent ARC studies further show that abstract reasoning benefits from visual
structure and modality-aware correction
\citep{zhang2026think,hu2026arc}. R-Qwen complements these approaches by
combining pretrained visual-language representations with explicit recursive
candidate refinement across heterogeneous symbolic and spatial tasks. Our contributions are:
\textbf{(1)} a task-general recursive search-and-refine framework built on a
pretrained Qwen backbone without architectural modifications;
\textbf{(2)} a unified candidate-refinement formulation for Sudoku, Maze, ARC,
and crosswords using task-specific serialization and constraint-preserving
projection;
\textbf{(3)} an adaptation of HSW~\citep{qasim2025cgar} to pretrained autoregressive models, emphasizing
the early, high-information refinement steps;
\textbf{(4)} an efficient parameter-efficient training strategy combining
LoRA, a recursion-depth curriculum, amortized candidate generation, and
per-instance early stopping, reducing training time and cost by more than
50\% in our experiments; and
\textbf{(5)} a systematic study of the training--inference recursion trade-off,
together with state-of-the-art results across eight benchmarks against prior
recursive models and much larger LLMs and VLMs.

\section{Related Work}
\label{sec:related_work}

\noindent\textbf{Recursive and autoregressive reasoning models.}
HRM and TRM increase effective computational depth through recursive refinement
and achieve strong performance on symbolic puzzles using small models trained
without chain-of-thought supervision
\citep{wang2025hrm,jolicoeurmartineau2025trm}. Subsequent work investigates
alternative recursive operators, including Mamba-based TRM variants
\citep{wang2026tiny}. Tiny Autoregressive Recursive Models examine whether
TRM-style refinement transfers directly to next-token prediction and find that
multi-step computation can help, but that directly transplanting latent
recursion into an autoregressive architecture does not consistently yield
reliable gains~\citep{rauba2026tiny}. R-Qwen instead formulates recursion as
prompt-conditioned editing of explicit structured candidates using a pretrained
backbone and parameter-efficient adaptation.

\noindent\textbf{Looped, fixed-point, and self-recursive models.}
LoopFormer repeatedly applies shared Transformer blocks with elastic
computational depth, while LOTUS refines latent tokens in a pretrained language
model under chain-of-thought supervision
\citep{jeddi2026loopformer,fan2026bridging}. Attractor Models replace finite
unrolling with a fixed-point solver in output-embedding space
\citep{feinashley2026solve}. Recursive Language Models instead decompose long
inputs and recursively call the language model on selected subproblems
\citep{zhang2026rlm}. In contrast, R-Qwen performs recursion over interpretable
task-level candidates, preserves task constraints after every update, and
requires neither a looped architecture nor recursive prompt decomposition.

\noindent\textbf{Visual abstract reasoning.}
VLSR combines visual and linguistic reasoning with modality-switch
self-correction, while VARC formulates ARC as image-to-image translation using
visual priors and test-time training
\citep{zhang2026think,hu2026arc}. These methods demonstrate the importance of
visual structure and iterative correction for ARC. R-Qwen addresses a
complementary problem: learning a task-general recursive refinement operator
that can be applied across visual, spatial, symbolic, and language-grounded
puzzles. A more detailed comparison with these research directions is provided
in Supplementary Section~\ref{SupplementaryMaterial}.

\begin{figure}
  \centering
        \includegraphics[width=0.48\textwidth]{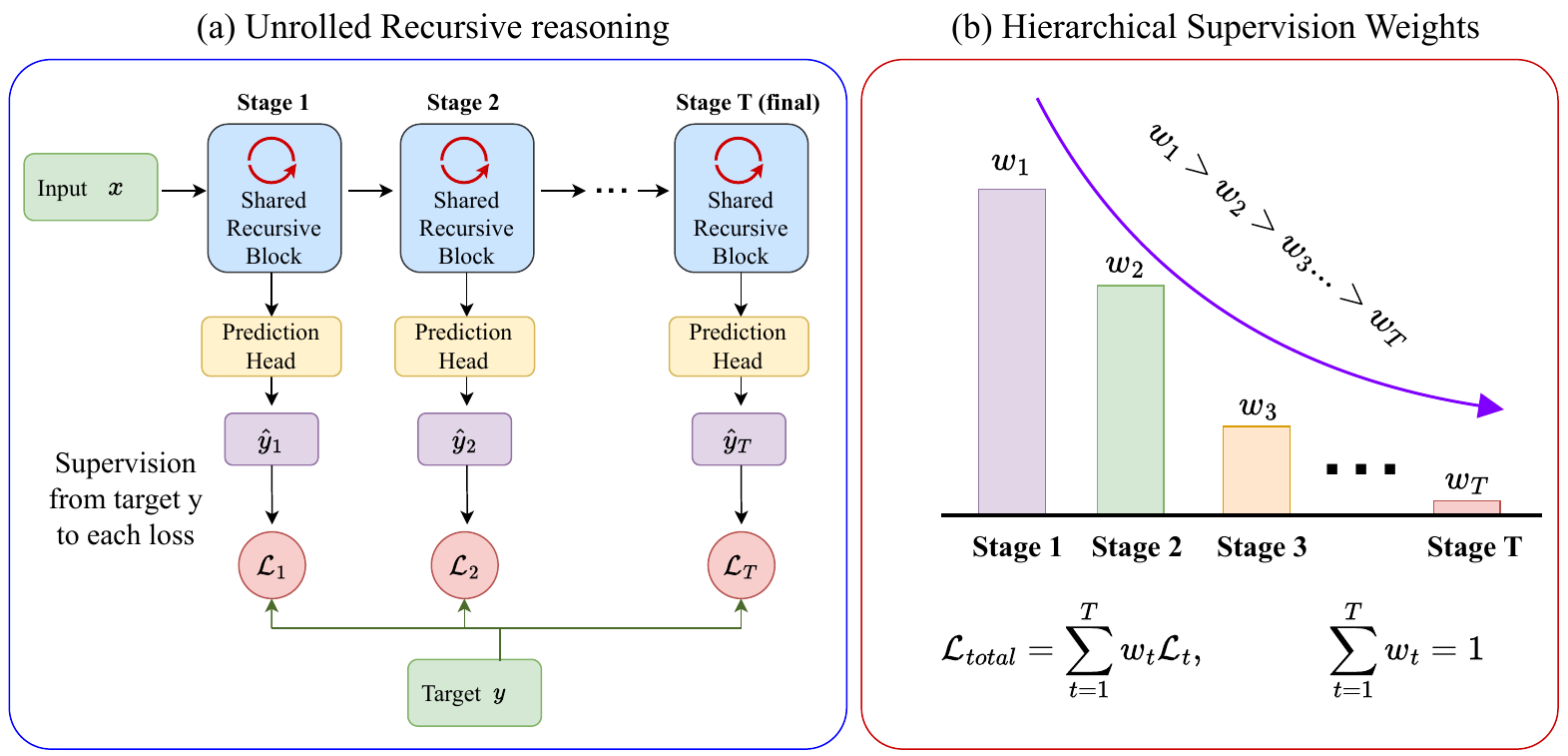}
        \caption{The proposed recursive reasoning framework with HSW.
(a) The recursive reasoning process is unrolled over $T$ stages using a shared recursive block. At each stage, the model produces an intermediate prediction $\hat{y}t$ through a prediction head, and each prediction is directly supervised by the target solution $y$, yielding a stage-wise loss $\mathcal{L}t$.
(b) HSW assigns larger weights to earlier stages and progressively smaller weights to later stages, i.e., $w_1 > w_2 > \cdots > w_T$. }
    \label{fig:hsw}
    \vspace{-15pt}
\end{figure}

\section{Methods}
\label{sec:method}

We cast a family of symbolic reasoning puzzles -- Sudoku, Maze,
ARC-AGI-1/2~\citep{chollet2019arc}, and CrossWordBench~\citep{leng2025crosswordbench}
-- as a unified \emph{iterative refinement}  problem, and fine-tune a single autoregressive LLM as a shared refinement operator. Our design follows the recursive-reasoning
and deep-supervision principle of the TRM~\citep{jolicoeurmartineau2025trm}, but departs from it in the substrate: rather
than the expensive training of a tiny non-autoregressive network that carries latent states
$(\bm{y},\bm{z})$ and a halting head, we use a pretrained autoregressive
backbone adapted with Low-Rank Adaptation (LoRA)~\citep{hu2022lora}, and we make
the refinement state \emph{explicit} as a serialized candidate that is fed back
into the model at every step. On top of this recurrence, we adapt HSW~\citep{qasim2025cgar},
replacing the uniform per-step weighting of standard deep supervision with a
principled exponential decay over refinement steps. The same training algorithm,
hyperparameters, and loss are used for every task; only the task-specific
serialization and constraint projection change.

\begin{figure*}
  \centering
        \includegraphics[width=\textwidth]{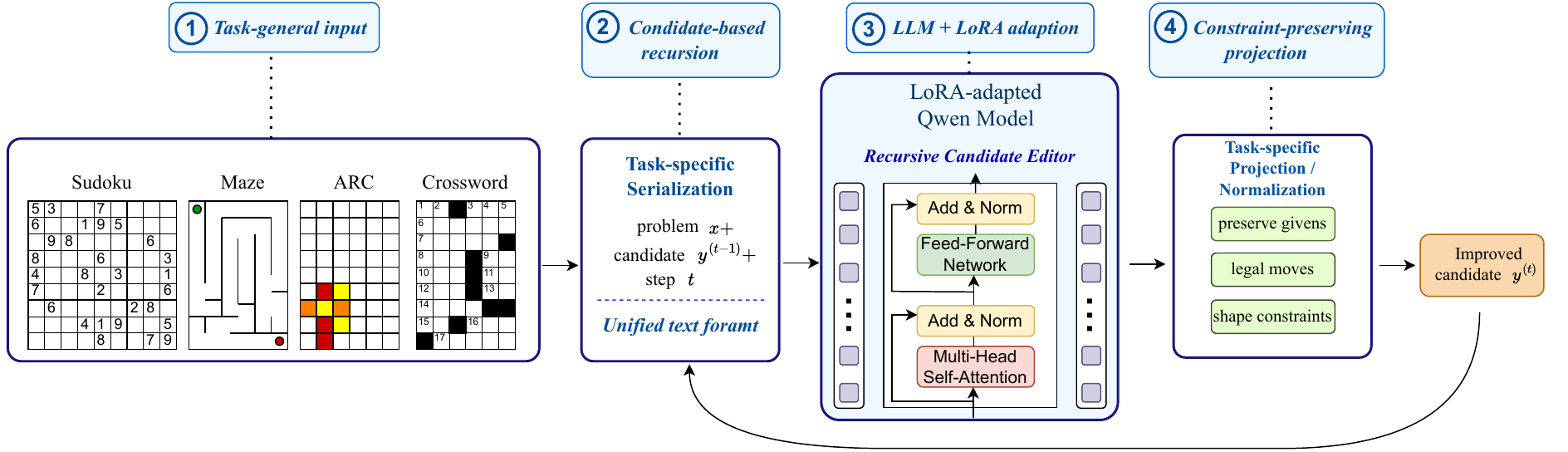}
        \caption{The proposed task-general recursive candidate refinement framework. 
Each structured reasoning task is serialized into a unified text format containing the original problem, the current candidate solution, and the refinement step. 
A LoRA-adapted autoregressive LLM generates an improved candidate, which is passed through a task-specific projection function to enforce output constraints. 
The refined candidate is then fed back into the model for the next recursion. 
During training, deep supervision is applied at every refinement step, and HSW assigns larger weights to earlier correction steps.}
    \label{fig:arch}
    \vspace{-15pt}
\end{figure*}

\noindent\textbf{Tasks as serialized constraint satisfaction.} We represent every task instance as a triple $(\bm{x},\,c,\,\bm{y}^{\ast})$.
Here $\bm{x}\in(\Sigma\cup\{\varnothing\})^{L}$ is the serialized
\emph{initial state} -- a length-$L$ string over a task alphabet $\Sigma$ in
which the symbol $\varnothing$ marks an unfilled position; $c$ is optional
auxiliary context (e.g. natural-language clues, demonstration pairs); and
$\bm{y}^{\ast}\in\Sigma^{L}$ is the serialized target solution. The length $L$
and alphabet $\Sigma$ are instance- and task-dependent. We write
$\mathcal{G}(\bm{x})=\{\,i : x_i\neq\varnothing\,\}$ for the set of
\emph{clamped} positions whose values are fixed by the instance and must be
preserved by any valid solution. We denote by
$\hat{\bm{y}}^{(t)}\in(\Sigma\cup\{\varnothing\})^{L}$ the candidate after
refinement step $t$ and initialise the recurrence with the instance itself,
\begin{equation}
\hat{\bm{y}}^{(0)} \;=\; \bm{x}.
\label{eq:init}
\end{equation}

This abstraction instantiates uniformly across our benchmarks
(Table~\ref{tab:tasks}): the alphabet, clamped set, and context differ, but the
operator, objective, and training loop are identical.

The model is a conditional language model $\pi_{\theta}$ whose trainable
parameters $\theta$ consist of frozen pretrained weights $\theta_0$ together
with low-rank adapters $\Delta\theta$ (see LoRa Section). A prompt
constructor $P(\bm{x},c,\hat{\bm{y}}^{(t-1)},t,N)$ renders a chat-formatted
input that contains (1) the instance $\bm{x}$ and context $c$, (2) the
current candidate $\hat{\bm{y}}^{(t-1)}$, and (3) the refinement index $t$ among $N$ steps, and instructs the model to return the improved solution as a serialized string while respecting the clamped positions. 

\noindent\textbf{Recursive candidate refinement.}
At step $t$ the model conditions on the current candidate and produces the next
one. Decoding is greedy, and the raw decoded string $\tilde{\bm{y}}^{(t)}$ is
passed through a deterministic, task-specific \emph{constraint projection}
$\Pi_{\bm{x}}(\cdot)$ before becoming the next state:

\begin{equation}
\begin{aligned}
\tilde{\bm{y}}^{(t)} 
&= \mathrm{Decode}_{\theta}\!\big(P(\bm{x},c,\hat{\bm{y}}^{(t-1)},t,N)\big),\\
\hat{\bm{y}}^{(t)}
&= \Pi_{\bm{x}}\!\big(\tilde{\bm{y}}^{(t)}\big).
\end{aligned}
\label{eq:refine}
\end{equation}

The projection enforces the hard structural constraints of the task and
guarantees a well-formed state regardless of generation quality. For each
position $i$,

\begin{equation}
\big[\Pi_{\bm{x}}(\tilde{\bm{y}})\big]_i =
\begin{cases}
x_i, & i \in \mathcal{G}(\bm{x}),\\[2pt]
\tilde{y}_i, & i \notin \mathcal{G}(\bm{x}),\ \tilde{y}_i \in \Sigma,\\[2pt]
\hat{y}^{(t-1)}_i, &  i \notin \mathcal{G}(\bm{x}) \text{ and } \tilde{y}_i \notin \Sigma.
\end{cases}
\label{eq:project}
\end{equation}

Here, positions in $\mathcal{G}(\bm{x})$ are clamped and remain fixed. Invalid, malformed, out-of-alphabet, or missing outputs are replaced by the previous prediction,
together with truncation or padding to the expected length $L$.
Equation~\eqref{eq:project} keeps the search on the feasible manifold (clamped
inputs such as Sudoku clues, maze walls, or crossword prefills are never
overwritten), makes the pipeline robust to malformed generations, and
relieves the model of having to re-emit the fixed part of the instance at every
step. Task-specific feasibility -- digit/colour/letter alphabets, grid shape,
and crossword intersection consistency -- is encoded entirely in
$\Sigma$, $L$, and $\Pi_{\bm{x}}$, leaving the learning algorithm unchanged.

\noindent
\textbf{Deep supervision over refinement steps.}
Following deep supervision~\citep{asadulaev2026latent} as instantiated for recursive reasoning in TRM, we attach a supervised loss to
\emph{every} refinement step rather than only to the final output. At step $t$
the model is teacher-forced to emit the \emph{full} target $\bm{y}^{\ast}$ from
the candidate-conditioned prompt, and we minimise the token-level cross-entropy over the target completion only (prompt tokens are masked):
\begin{equation}
\ell^{(t)}(\theta) \;=\;
-\frac{1}{|\bm{y}^{\ast}|}\sum_{i=1}^{|\bm{y}^{\ast}|}
\log \pi_{\theta}\!\Big(y^{\ast}_{i}\;\Big|\;
P\big(\bm{x},c,\hat{\bm{y}}^{(t-1)},t,N\big),\, \bm{y}^{\ast}_{<i}\Big).
\label{eq:st-loss}
\end{equation}
The per-sample training objective aggregates these step losses with weights
$\{w_t\}_{t=1}^{N}$:
\begin{equation}
\mathcal{L}(\theta)\;=\;\sum_{t=1}^{N} w_t\,\ell^{(t)}(\theta).
\label{eq:objective}
\end{equation}
Standard deep supervision uses uniform weights $w_t=1/N$; in
Eq~\ref{eq:hsw} we replace these with a recursion-aware decay. Because the
target is the complete solution at every step, the model is trained to map an
\emph{arbitrary partially-correct candidate} back to the solution, which is
exactly the operator iterated at inference time. Gradients flow only through the
current step; following the truncated-recurrence treatment of recursive
models~\citep{jolicoeurmartineau2025trm, behzadian2026upi}, candidate states produced by decoding are detached and treated as fixed inputs to subsequent steps.

\noindent
\textbf{Supervision-depth curriculum, amortised generation, and early stopping.}
Difficulty is highly non-uniform within and across tasks:
most instances are solved in fewer steps than the maximum
recursion horizon. Applying the full refinement depth from
the first epoch therefore wastes computation and may
over-train the model on easier instances
\cite{qasim2025cgar}. We consequently introduce a
curriculum over the number of supervised refinement steps
\cite{hammoud2025train}. Let $e\in\{1,\ldots,E\}$ index
the training epochs, and let
$1=e_1<e_2<\cdots<e_S\leq E$ denote the epochs at which
the recursion horizon is increased. Defining $e_{S+1}=E+1$,
we set the active horizon $N\equiv N_{\mathrm{sup}}(e)$ as:


\begin{equation}
N_{\mathrm{sup}}(e)
=
N_k
\quad
\text{for }
e_k\leq e<e_{k+1},
\qquad
k=1,\ldots,S,
\label{eq:curr}
\end{equation}
\noindent
where
$N_1\leq N_2\leq\cdots\leq N_S\leq N_{\max}$.
For example, choosing
$(e_1,e_2,e_3)=(1,4,7)$ and
$(N_1,N_2,N_3)=(4,8,10)$ yields four refinement steps
during epochs $1$--$3$, eight steps during epochs $4$--$6$,
and ten steps from epoch $7$ onward. Shallow recursion early
in training stabilises the adapters and acts as a regulariser,
whereas deeper recursion later exposes the model to the
longer refinement trajectories required by difficult
instances. This resembles the Progressive Depth Curriculum
of CGAR \cite{qasim2025cgar}, except that we schedule the
outer refinement horizon of an autoregressive operator rather
than the inner $(n,T)$ recursion of a small network.
Two further mechanisms reduce the cost of the inner loop without materially changing the learning signal:
\begin{itemize}
  \item \textbf{Amortised candidate generation.} Autoregressive decoding is the
  dominant per-step cost, especially for the larger grids of ARC and
  CrossWordBench, where the decode budget scales with $L$. We regenerate the
  candidate only on a subset of steps
  $\mathcal{S}_{\mathrm{gen}}=\{t : t\equiv 0 \pmod{m}\}\cup\{N\}$ (every $m$th
  step, and always at the final step); on the remaining steps the candidate is
  carried forward, $\hat{\bm{y}}^{(t)}=\hat{\bm{y}}^{(t-1)}$. The supervised
  loss of Eq.~\eqref{eq:st-loss} is still applied at \emph{every} step, so the
  operator continues to receive deep supervision while expensive generation is
  amortised over $m$ updates.
  \item \textbf{Per-instance early stopping.} We maintain an active set
  $\mathcal{A}_t=\{\,n : \hat{\bm{y}}_n^{(t-1)}\neq \bm{y}_n^{\ast}\,\}$ and only
  compute losses and decode for unsolved instances; once an instance reaches its
  target it is removed, and a batch terminates as soon as
  $\mathcal{A}_t=\varnothing$. This concentrates compute on instances that still
  carry a learning signal.
\end{itemize}

\noindent
\textbf{Hierarchical Supervision Weighting (HSW).}
We observed that TRM uses unusually large weight decay during optimization. This can be explained as follows: uniform weighting $w_t=1/N$ in Eq.~\eqref{eq:objective} implicitly assumes that
every refinement step is equally informative. In recursive refinement this is
not the case: the largest corrections occur in the first few steps, after which
the candidate is close to the solution and successive steps contribute
exponentially diminishing -- yet increasingly noisy -- gradients. Treating all
steps equally therefore over-weights low-information late steps and inflates the
variance of the stochastic gradient. Motivated by this intuition, and following the empirical observation of step-wise gradient-magnitude decay underlying HSW~\citep{qasim2025cgar}, we adopt the following weighting:

\begin{align}
w_t
&= \frac{\lambda^{\,t-1}}{Z_{\lambda}},
\notag\\
Z_{\lambda}
&= \sum_{s=1}^{N}\lambda^{\,s-1}
= \frac{1-\lambda^{N}}{1-\lambda},
\notag\\
&\qquad \lambda\in(0,1].
\label{eq:hsw}
\end{align}
\noindent
The decay rate $\lambda$ interpolates between supervision concentrated almost
entirely on the first refinement step ($\lambda\!\to\!0$) and standard uniform
deep supervision ($\lambda=1$, for which $w_t=1/N$). Since
$w_1>w_2>\cdots>w_N$ for $\lambda\in(0,1)$, earlier refinement steps receive
larger supervision weights than later ones. Substituting
Eq.~\eqref{eq:hsw} into Eq.~\eqref{eq:objective} yields the HSW objective
\begin{equation}
\mathcal{L}_{\mathrm{HSW}}(\theta)\;=\;
\frac{1}{Z_{\lambda}}\sum_{t=1}^{N}\lambda^{\,t-1}\,\ell^{(t)}(\theta),
\label{eq:hsw-objective}
\end{equation}
which focuses gradient signal on the early, high-information refinement steps
while preserving a small contribution from later steps so that long-horizon
refinement is still learned. Because the normaliser $Z_{\lambda}$ is recomputed
for the current horizon $N=N_{\mathrm{sup}}(e)$, HSW composes cleanly with the supervision-depth curriculum of Eq.~\eqref{eq:curr}: the weights always form a valid distribution over the active number of steps. HSW is task-agnostic,
adds no parameters, and incurs no extra forward or backward passes; CGAR reports
that the resulting variance reduction improves the signal-to-noise ratio of
stochastic gradients and accelerates convergence at matched
accuracy~\citep{qasim2025cgar}.


\paragraph{Realisation under per-step updates.}
Equation~\eqref{eq:hsw-objective} describes a single aggregated update per
instance. Our training loop instead performs one optimiser step per refinement
step, so we realise HSW by scaling each step loss before its backward pass,
$\ell^{(t)}\!\leftarrow\! w_t\,\ell^{(t)}$. Under per-step SGD-style updates the
weight $w_t$ acts as a step-dependent modulation of the effective gradient norm
(equivalently, a per-step learning rate), preserving the intended emphasis on
early steps; accumulating the weighted losses and applying a single update per
batch recovers Eq.~\eqref{eq:hsw-objective} exactly and is a drop-in
alternative.

\noindent
\textbf{Parameter-efficient adaptation.}
We keep the pretrained backbone frozen and inject trainable LoRA into the attention projections. For a target weight
matrix $W_0\in\mathbb{R}^{d\times k}$, the adapted forward map is

\begin{align}
W
&= W_0 + \frac{\alpha}{r} B A,
\notag\\
&\qquad B\in\mathbb{R}^{d\times r},\ A\in\mathbb{R}^{r\times k},\ r\ll \min(d,k).
\label{eq}
\end{align}

\noindent
with rank $r$, scaling $\alpha$, and adapter dropout on the input of $A$. Only
$\{A,B\}$ (and no biases) are trained, so the trainable-parameter count is a
small fraction of the backbone, and a single set of adapters is shared across
all tasks. Optionally the frozen backbone is loaded in 4-bit NF4 with double
quantisation (QLoRA~\citep{dettmers2023qlora}) to reduce memory. Optimisation
uses SGD with gradient clipping, and a cosine
learning-rate schedule with linear warmup; the total number of optimiser steps
is computed from the curriculum in Eq.~\eqref{eq:curr} as
$\sum_{e=1}^{E}\lceil |\mathcal{D}_e|/B\rceil\, N_{\mathrm{sup}}(e)$, where $B$
is the batch size and $\mathcal{D}_e$ the (possibly sharded) training set used
at epoch $e$.
 
\noindent\textbf{Recursive inference and metrics.}
At test time we apply the learned operator for a fixed horizon of $K$ steps.
Starting from $\hat{\bm{y}}^{(0)}=\bm{x}$, we iterate the
generation--projection update of Eq.~\eqref{eq:refine} for $t=1,\dots,K$ and
return $\hat{\bm{y}}^{(K)}$. The same constraint projection $\Pi_{\bm{x}}$ used
in training is applied at every inference step, so all intermediate states are
feasible and the clamped inputs are preserved by construction. For each task we
report position-level accuracy (fraction of correct cells/tokens) and
exact-match accuracy (whole-solution correctness), both at the final step and as
a function of $t$; the latter exposes the convergence behaviour of the
refinement process and lets us compare the number of steps each task requires.
Unlike inference-time recursive schemes that recurse over the
\emph{prompt}~\citep{zhang2026rlm, venkatraman2025recursive} or looped models with learned elastic
depth~\citep{jeddi2026loopformer}, our recurrence is over an explicit
task-level candidate and the per-step budget is fixed and interpretable across
all eight benchmarks.

\section{Experiments}
\textbf{Datasets.}
We evaluate the proposed recursive candidate refinement framework on a diverse set of structured reasoning benchmarks, including Sudoku, Maze, ARC-AGI-1, ARC-AGI-2, and CrossWordBench. These datasets cover different forms of reasoning: constraint satisfaction, path planning, abstract visual transformation, and language-grounded grid completion. This allows us to study whether recursive refinement is useful across multiple structured domains rather than being specialized to a single task. Detailed descriptions of each dataset are provided in the Appendix.





\begin{table}[t]
\centering
\caption{Sudoku, Maze, ARC-1, and ARC-2 comparisons. LLM-based results are from the ARC-AGI leader-
board. HRM, TRM, Loopformer and VARC are trained from scratch. $\dagger$ denotes the number of trainable parameters.}
\label{tab:puzzle_chal}
\resizebox{\columnwidth}{!}{
\begin{tabular}{lccccc}
\toprule
\textbf{Method} & \textbf{\# Params} & \textbf{Sudoku} & \textbf{Maze} & \textbf{ARC-1} & \textbf{ARC-2} \\
\midrule
\rowcolor{gray!15}\multicolumn{6}{l}{\textit{large language models (LLMs)}} \\
Deepseek R1            & 671B & 0.0  & 0.0  & 15.8 & 1.3  \\
Claude 3.7             & N/A    & 0.0  & 0.0  & 28.6 & 0.7  \\
o3-mini-high           & N/A    & 0.0  & 0.0  & 34.5 & 3.0  \\
Gemini 2.5 Pro 32K     & N/A    & --   & --   & 37.0 & 4.9  \\
GPT-5                  & N/A    & 58.6   & 65.1   & 44.0 & 1.9  \\
Grok-4-thinking        & 1.7T & 61.8   & 73.5   & 66.7 & 16.0 \\
Bespoke (Grok-4)       & 1.7T & --   & --   & 79.6 & 29.4 \\
\midrule
\rowcolor{gray!15}\multicolumn{6}{l}{\textit{vision models}} \\
VARC (CVPR 2026)                   & 73M  & - & - & 60.4 & 11.1  \\
\midrule
\rowcolor{gray!15}\multicolumn{6}{l}{\textit{recursive models}} \\
HRM                    & 27M  & 55.0 & 74.5 & 40.3 & 5.0  \\
TRM        & 7M   & 74.7 & 85.3 & 44.6 & 7.8  \\
LoopFormer (ICLR 2026)         & 278M & 67.3 & 78.5 & 39.1 & 1.9 \\
\rowcolor{green!3} Ours      & 27B (12M $^\dagger$) & 88.5 & 93.1 & 82.1 & 33.5 \\
\bottomrule
\end{tabular}
}
\footnotetext{TRM-MLP uses 5M parameters on Sudoku and 19M parameters on Maze; ARC-AGI uses 19M parameters.}
\end{table}

\begin{table}[!t]
  \caption{Comparison with vision-language co-reasoning methods on ARC-style abstract reasoning benchmarks.}
  \centering
  \resizebox{\columnwidth}{!}{
  \normalsize
  \begin{tabular}{l|c|c|c}
    \toprule
    Models  & ARC-AGI & BARC-100 & Re-ARC\\
    \midrule
    GPT-4o & 8.2 & 28.0 & 10.0 \\
    Gemini-2.5-Pro & 35.0 & 56.0 & 30.0 \\
    o4-mini & 42.2 & 59.0 & 36.0 \\
    Qwen3-VL-235B & 20.2 & 52.0 & 20.0 \\
    LoopFormer (ICLR 2026) & 13.4 & 33.0 & 14.0 \\
    VLSR (CVPR 2026) & 22.0 & 52.0 & 21.0  \\
   \rowcolor{green!3} Ours & 47.8 & 66.0 & 41.0 \\
    \bottomrule
  \end{tabular}}
  \vspace{-15pt}
  \label{tab:puzzle_vl}
\end{table}

\begin{table}[t]
\centering
\caption{Comparison of various LLMs and LVLMs on CrossWordBench English set across two difficulty levels. Results are in percentages over 100 samples for both difficulties. $\Delta$ indicates that the model is a reasoning model.}
\label{tab:crosswordbench}
\small
\resizebox{\columnwidth}{!}{%
\begin{tabular}{lcccccc}
\toprule
\multirow{2}{*}{\textbf{Models}}
& \multicolumn{3}{c}{\textbf{7$\times$7}}
& \multicolumn{3}{c}{\textbf{14$\times$14}} \\
\cmidrule(lr){2-4} \cmidrule(lr){5-7}
& \textbf{WCR} & \textbf{LCR} & \textbf{ICR}
& \textbf{WCR} & \textbf{LCR} & \textbf{ICR} \\

\midrule
\rowcolor{gray!15}\multicolumn{7}{l}{\textit{Proprietary LLMs}} \\
o3-mini-high
& $58.7$ & $68.4$ & \cellcolor{green!20}$89.1$
& $44.5$ & $52.0$ & \cellcolor{green!20}$51.2$ \\
Claude-3-7-Sonnet$^\Delta$
& $61.7$ & $71.2$ & $75.4$
& $49.2$ & $54.2$ & $43.1$ \\
Claude-3-7-Sonnet
& $48.2$ & $57.4$ & $47.2$
& $44.6$ & $48.5$ & $32.1$ \\
GPT-4o-2024-11-20
& $41.0$ & $47.2$ & $28.8$
& $33.8$ & $36.9$ & $19.6$ \\
Gemini 2.0 Pro Exp
& $46.0$ & $52.5$ & $38.8$
& $42.5$ & $45.7$ & $28.9$ \\
Gemini 2.0 Flash
& $30.1$ & $31.8$ & $25.5$
& $28.0$ & $29.8$ & $19.8$ \\

\midrule
\rowcolor{gray!15}\multicolumn{7}{l}{\textit{Open-Weight LLMs}} \\
Llama-3.1-405B-Instruct
& $16.1$ & $35.9$ & $24.3$
& $35.5$ & $39.0$ & $22.2$ \\
DeepSeek-R1
& $64.6$ & $70.7$ & $67.8$
& $47.2$ & $50.7$ & $35.6$ \\
DeepSeek-V3
& $30.3$ & $36.9$ & $18.6$
& $29.0$ & $33.5$ & $14.5$ \\
R1-Distill-Llama-70B
& $38.7$ & $44.8$ & $34.7$
& $28.5$ & $31.9$ & $16.1$ \\
Llama-3.3-70B-Instruct
& $30.3$ & $37.1$ & $20.6$
& $28.0$ & $34.0$ & $17.3$ \\
QwQ-32B
& $34.7$ & $44.5$ & $51.8$
& $25.4$ & $30.7$ & $18.9$ \\
Open-Reasoner-Zero-32B
& $13.9$ & $20.4$ & $18.4$
& $14.6$ & $19.9$ & $9.5$ \\
Phi-4
& $12.2$ & $19.4$ & $11.3$
& $14.0$ & $20.0$ & $8.5$ \\
\midrule
Ours
& \cellcolor{green!20}$65.7$ & \cellcolor{green!20}$72.8$ & $75.1$
& \cellcolor{green!20}$51.3$ & \cellcolor{green!20}$55.0$ & $45.2$ \\
\bottomrule
\end{tabular}}
\vspace{-15pt}
\end{table}

\noindent 
\textbf{Experimental Setup.}
We fine-tune a pretrained Qwen3.5-27B backbone using LoRA adaptation while keeping the backbone weights frozen. Unless otherwise stated, LoRA is applied to all linear layers of the attention blocks. The model is trained with supervised cross-entropy on the target completion, masking prompt tokens from the loss, and optionally supports 4-bit QLoRA for memory-efficient training.

For all tasks, the model follows the recursive candidate refinement framework described in Section~3. At refinement step $t$, it receives the original problem $\mathbf{x}$, the current candidate $\hat{\mathbf{y}}^{(t-1)}$, and the refinement index $t$, predicts an improved candidate, and applies a task-specific projection for the next recursion step. All benchmarks use the same recursive training framework and hyperparameters, differing only in task-specific serialization and constraint projection. Implementation details, optimization settings, curriculum schedule, and training hyperparameters are provided in the Appendix.
\begin{figure*}[t]
    \centering

    \includegraphics[width=0.32\textwidth]
        {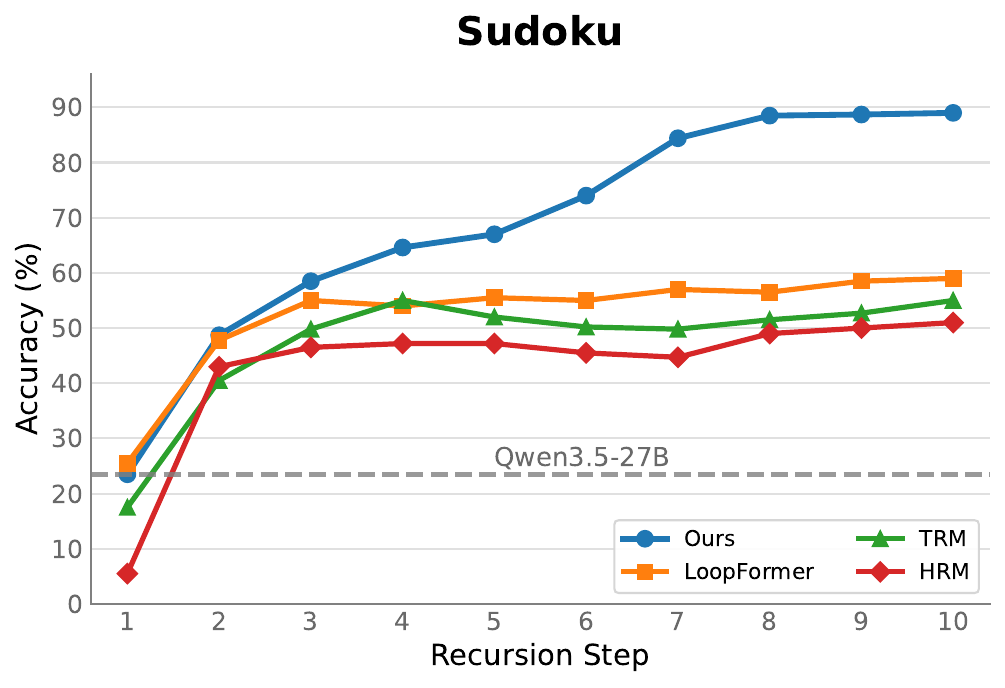}
    \hfill
    \includegraphics[width=0.32\textwidth]
        {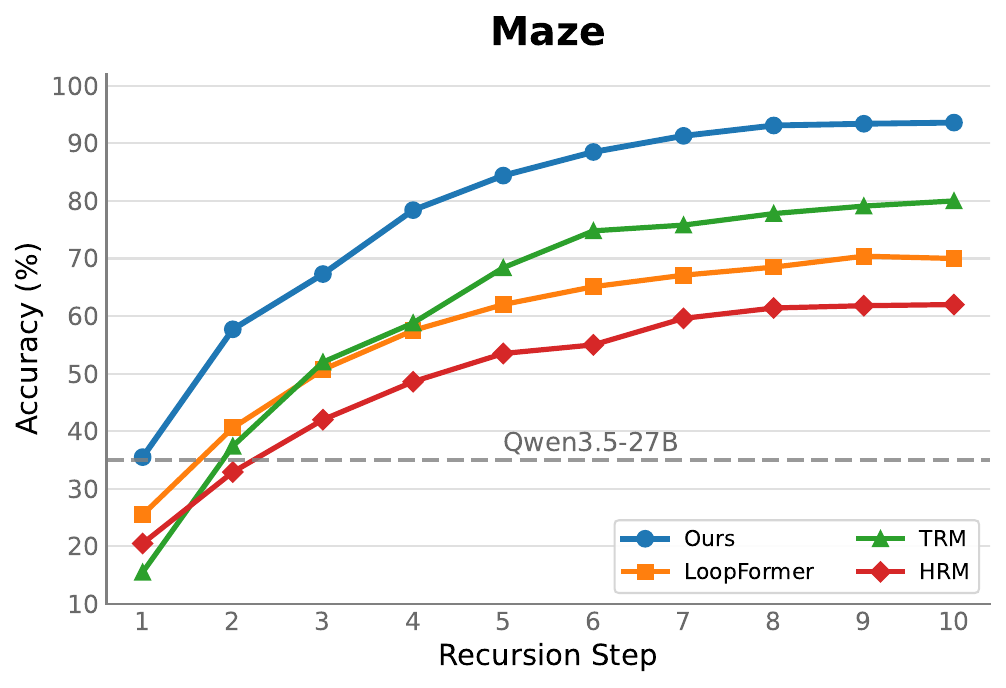}
    \hfill
    \includegraphics[width=0.32\textwidth]
        {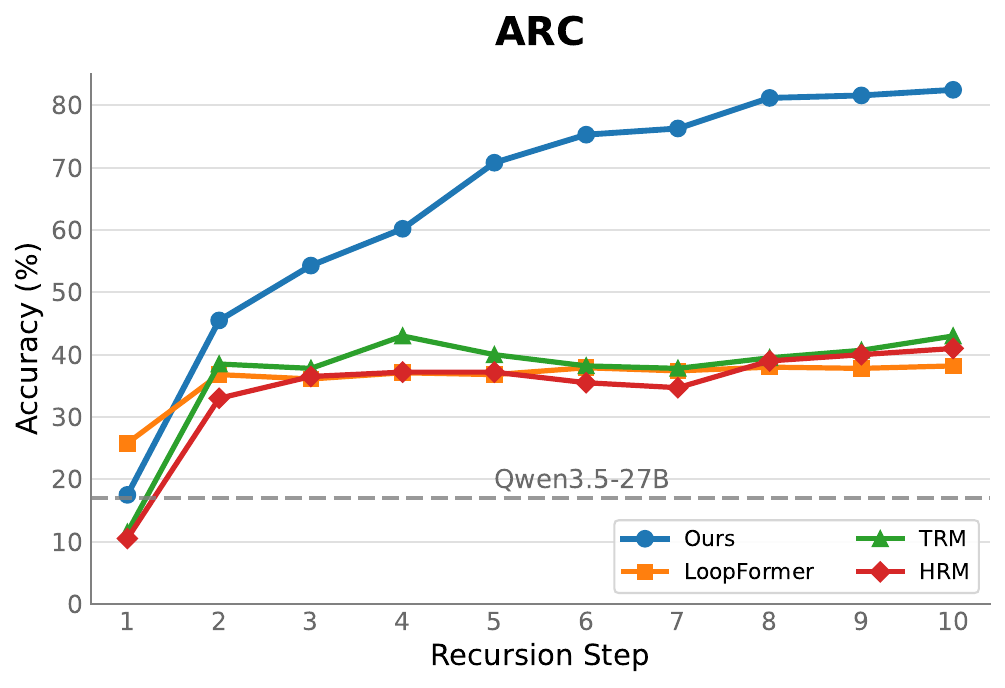}

    \caption{Accuracy comparison across recursion steps on three benchmarks. Our method achieves substantially higher accuracy than LoopFormer, TRM, and HRM as the aggregation step increases. The dashed gray line indicates the Qwen3.5-27B baseline.}
    \label{fig:aggregation_all}
\end{figure*}

\noindent
\textbf{Challenging Puzzle Tasks.}
Table~\ref{tab:puzzle_chal} presents a system-level comparison on four challenging symbolic-reasoning benchmarks: Sudoku, Maze, ARC-1, and ARC-2. We compare our method against three families of approaches: LLMs, VLMs, and recursive reasoning models. The LLM-based ARC results are taken from the ARC-AGI leaderboard, while HRM, TRM, LoopFormer, VARC, and our method are evaluated in the setting where models are trained or adapted on the corresponding benchmark data.

Our method compares favorably with strong frontier LLMs. Although recent LLMs such as GPT-5 and Grok-4-thinking achieve non-trivial performance on Sudoku, Maze, and ARC, they rely on very large pretrained models with hundreds of billions to trillions of parameters. In contrast, our method achieves higher accuracy across all four benchmarks. Compared with Grok-4-thinking, our method improves performance by 26.7\%, 19.6\%, 15.4\%, and 17.5\% on Sudoku, Maze, ARC-1, and ARC-2, respectively.
In the controlled comparison with recursive reasoning models, our method substantially outperforms HRM, TRM, and LoopFormer. Compared with TRM, which is the strongest recursive baseline on Sudoku and Maze, our method achieves improvements of 13.8\% on Sudoku, 7.8\% on Maze, 37.5\% on ARC-1, and 25.7\% on ARC-2.
Our method also improves over the vision-based VARC model on ARC. VARC demonstrates that ARC can benefit from visual inductive biases, achieving 60.4 on ARC-1 and 11.1 on ARC-2. Our method further increases these results to 82.1 and 33.5, respectively. This suggests that combining strong pretrained representations with recursive refinement provides a more effective mechanism for solving both visual abstraction tasks and structured symbolic-reasoning problems.

Overall, our method achieves the highest accuracy across all four benchmarks while fine-tuning only 12M task-specific parameters on top of a 27B backbone. These results show that recursive refinement with pretrained representations is effective for structured symbolic reasoning.

\begin{table}[t]
\centering
\caption{HSW vs. halting act (HA) in convergence speed. Training cost is computed at \$3.99/hour for TRM with one H100 GPU and \$12.36/hour for Qwen with four H100 GPUs.}
\label{tab:HSW_comparison}
\small
\resizebox{\columnwidth}{!}{%
\begin{tabular}{lcccc}
\toprule
\textbf{Method} 
& \textbf{Exact Acc. (\%)} 
& \textbf{Time (h)} 
& \textbf{Speedup} 
& \textbf{Cost} \\
\midrule
TRM  & 74.7 & 23.4 & $1.0\times$  & \$93.37 \\
+HA  & 74.1 & 18.5 & $1.26\times$ & \$73.82 \\
+HSW & 75.3 & 11.2 & $2.09\times$ & \$44.69 \\
\midrule
Qwen & 85.1 & 45.8 & $1.0\times$  & \$566.09 \\
+HA  & 84.6 & 39.1 & $1.17\times$ & \$483.28 \\
+HSW & 88.5 & 22.5 & $2.04\times$ & \$278.10 \\
\bottomrule
\end{tabular}}
\vspace{-15pt}
\end{table}

\textbf{Visual-Language Puzzles.}
Table~\ref{tab:puzzle_vl} reports Pass@1 accuracy (\%) on three abstract reasoning benchmarks: ARC-AGI, BARC-100, and Re-ARC. Our method achieves the best performance across all benchmarks, reaching 47.8\% on ARC-AGI, 66.0\% on BARC-100, and 41.0\% on Re-ARC.
Compared with the strongest LLM baseline, o4-mini, our method improves accuracy by 5.6\%, 7.0\%, and 5.0\% on ARC-AGI, BARC-100, and Re-ARC, respectively. Compared with the VLSR baseline \cite{zhang2026think}, our method further improves accuracy by 25.8\%, 14.0\%, and 20.0\%. These gains show that our approach provides consistent improvements over both strong general-purpose models and prior visual-language co-reasoning methods. Importantly, our method is orthogonal to vision-language synergy, suggesting that additional gains may be achieved by incorporating such synergy into R-Qwen.
Compared with LoopFormer, our method improves accuracy by 34.4\% on ARC-AGI, 33.0 \% on BARC-100, and 27.0\% on Re-ARC. Overall, the results demonstrate that combining visual-language co-reasoning with effective self-correction leads to stronger abstract reasoning performance across diverse ARC-style benchmarks.

\begin{figure}[t]
    \centering
    \includegraphics[width=\columnwidth]{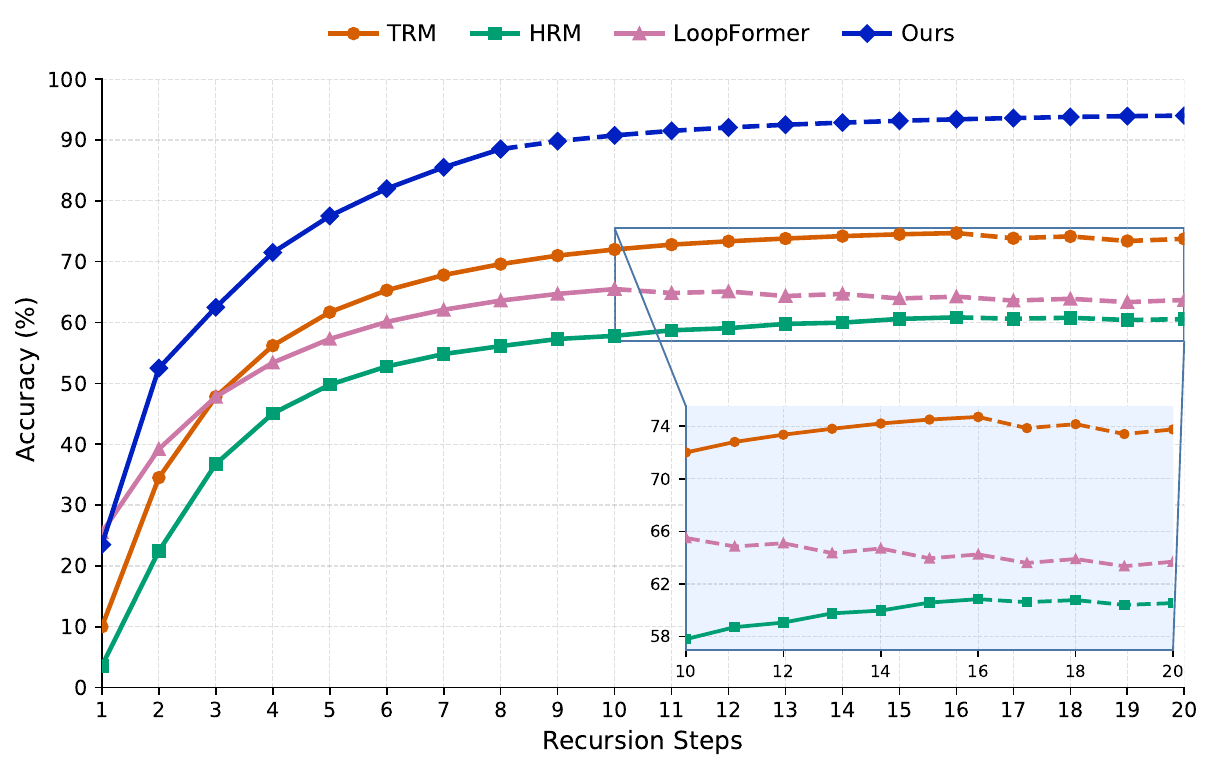}
    \caption{Exact accuracy across recursion steps. The right panel magnifies the later recursion steps to highlight performance fluctuations after steps 10 and 16.}
    \label{fig:recursion_step}
    \vspace{-15pt}
\end{figure}

\textbf{Crossword Puzzle Task.}
Table~\ref{tab:crosswordbench} reports results on the CrossWordBench English set. Following the original benchmark protocol, we evaluate models on two grid sizes, $7\times7$ and $14\times14$, using three metrics: Word Coverage Rate (WCR), Letter Coverage Rate (LCR), and Intersection Consistency Rate (ICR). WCR and LCR measure word-level and letter-level accuracy, while ICR evaluates whether the predicted words satisfy crossing-letter constraints.

\begin{figure}[t]
    \centering
    \includegraphics[width=\columnwidth]{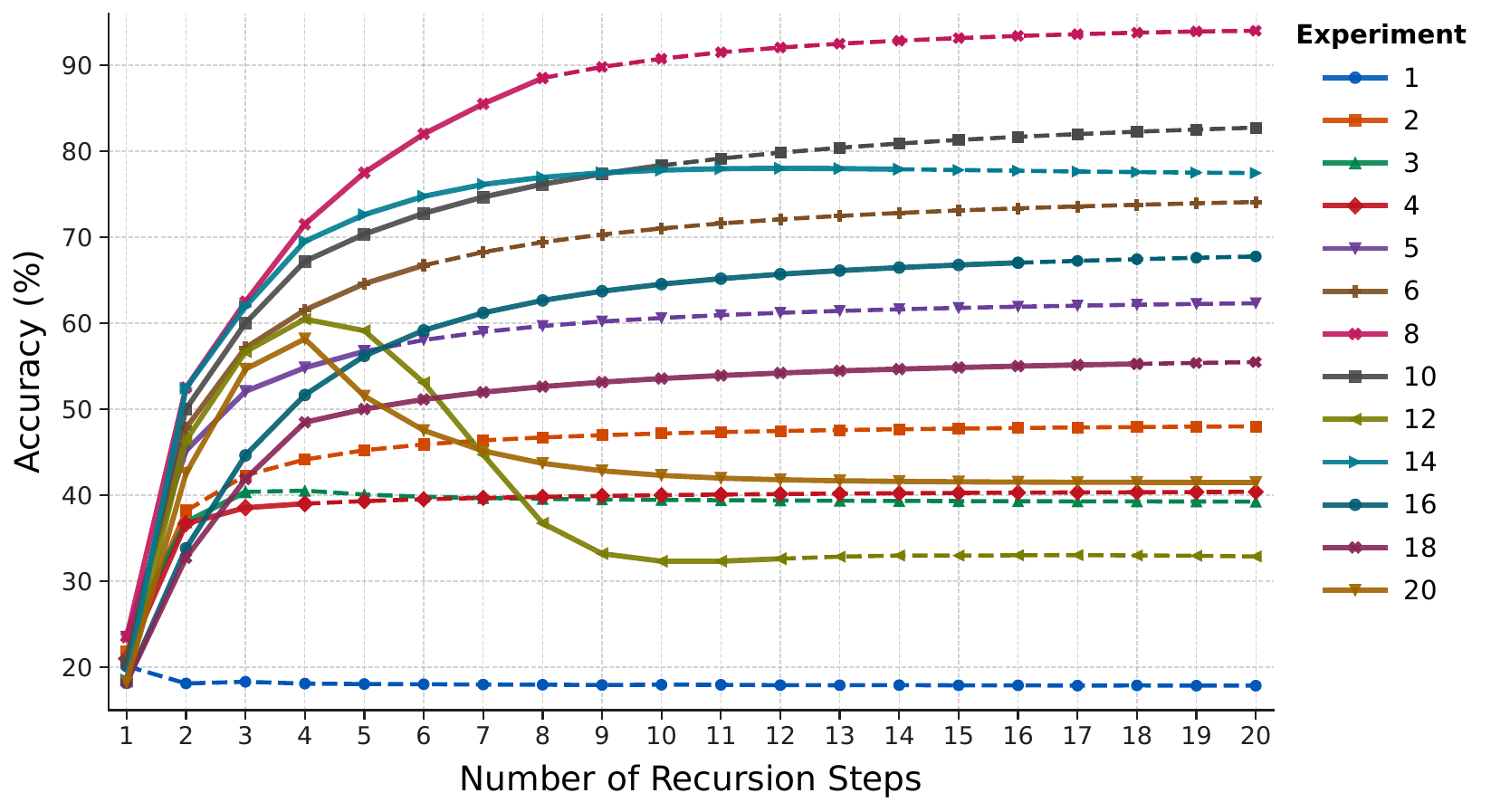}
    \caption{Exact accuracy across recursion steps. The right panel indicates the number of recursion steps used during training. Each curve is shown as a solid line up to the training recursion number and as a dashed line beyond that depth.}
    \label{fig:train_eval}
    \vspace{-15pt}
\end{figure}

Our method achieves the strongest overall accuracy on WCR and LCR across both grid sizes. On the $7\times7$ setting, our method reaches 65.7\% WCR and 72.8\% LCR, improving over the strongest previous results by 1.1\% and 1.6\%, respectively. On the more challenging $14\times14$ setting, our method obtains 51.3\% WCR and 55.0\% LCR, improving the best baseline by 2.1\% and 0.8\%, respectively.
Compared with the strongest open-weight baseline, DeepSeek-R1, our method improves WCR by 1.1\% and LCR by 2.1\% on $7\times7$ grids, and by 4.1\% and 4.3\% on $14\times14$ grids. These gains indicate that our method better preserves both word-level correctness and fine-grained letter-level accuracy, especially as puzzle complexity increases.
For ICR, o3-mini-high remains the strongest model, achieving 89.1\% on $7\times7$ and 51.2\% on $14\times14$. Nevertheless, our method remains competitive, achieving 75.1\% and 45.2\% ICR on the two settings. Overall, the results show that our method improves crossword-solving accuracy across grid sizes while maintaining strong consistency under intersection constraints.

\subsection{Ablation studies}
\textbf{Number of Recursive Steps:} 
To determine the optimal number of recursive steps, we train separate models with different recursion depths and evaluate them on Sudoku, Maze, and ARC. As shown in Figure~\ref{fig:aggregation_all}, performance consistently improves as the number of steps increases. The largest gain is observed from step 1 to step 2, while the improvement gradually decreases with deeper recursion. After around 8 steps, the performance nearly saturates and further gains become marginal. Therefore, we use 8 recursive steps in our main experiments as a good trade-off between accuracy and efficiency. This ablation also shows that our method benefits more effectively from recursive refinement compared with LoopFormer, TRM, and HRM.

\noindent \textbf{Training Efficiency:} 
We evaluate the effect of HSW on both accuracy and training efficiency, compared with the standard model and the Halting Act (HA) method. As shown in Table~\ref{tab:HSW_comparison}, HSW consistently improves convergence speed while preserving or improving accuracy. For TRM, HSW reduces training time from 23.4 hours to 11.2 hours, achieving a $2.09\times$ speedup and slightly improving exact accuracy 0.6\%. Similarly, for Qwen, HSW reduces training time from 45.8 hours to 22.5 hours, achieving a $2.04\times$ speedup while improving accuracy 3.4\%. Compared with HA, HSW provides both higher accuracy and faster convergence.

\noindent \textbf{Evaluation Recursion Steps:} 
We further study the effect of increasing the number of recursion steps at evaluation time. As shown in Figure~\ref{fig:recursion_step}, our method consistently improves with additional refinement steps. In contrast, LoopFormer begins to fluctuate after around step 10, while TRM and HRM show similar fluctuation and slight performance drop after around step 16. R-Qwen remains stable and continues to improve when evaluated beyond its
training recursion horizon. Although HSW may contribute to this behavior by
emphasizing earlier, more informative refinement steps, the comparison in
Figure~\ref{fig:recursion_step} does not isolate its effect. Determining whether this
inference-time robustness is specifically attributable to HSW requires a
controlled ablation against uniform deep supervision, which is a subject of future work.


\noindent \textbf{Training--Evaluation Trade-off:}
This ablation study investigates the impact of using different numbers of recursion steps during both training and inference. As shown in Figure~\ref{fig:train_eval}, simply applying recursive refinement to a pretrained VLM at inference time (Experiment 1) does not improve performance; instead, accuracy remains nearly unchanged and even degrades slightly with additional recursion. This observation is consistent with recent studies showing that LLMs are generally unable to improve their own reasoning through repeated self-revision without additional supervision or external feedback~\cite{huang2024large}. In contrast, recursion-aware training enables the model to learn an effective iterative refinement policy. 

Figure~\ref{fig:train_eval} shows that longer training horizons do not monotonically improve performance: moderate recursion helps, whereas 20 steps reduces accuracy. Similar behavior has been reported in recursive and looped reasoning models~\cite{jolicoeurmartineau2025trm,jeddi2026loopformer}. Among the evaluated settings, training with eight recursion steps achieves the best overall performance, attaining the highest accuracy while remaining stable beyond its training horizon. Moreover, applying a few additional recursion steps at inference time yields a small but consistent improvement, suggesting that the learned refinement policy generalizes to trajectories longer than those encountered during training.

Models trained with longer horizons, particularly 12 or 20, exhibit an apparent overthinking effect: accuracy initially increases, reaches a peak, and then declines as recursion continues. This pattern is consistent with other iterative reasoning systems~\cite{ren2026your,bansal2022end}. Across all settings, most improvement occurs within the first few recursion steps, after which the gains diminish and saturate. These results indicate that successful recursive reasoning depends on learning a stable refinement policy rather than simply increasing the number of iterations, requiring a careful balance between the training and inference horizons.

\section{Conclusions}
We introduce R-Qwen, a recursive reasoning framework that brings iterative refinement to a pretrained language model backbone. Unlike prior recursive reasoning models trained from scratch for individual tasks, R-Qwen combines the structured search-and-correction behavior of recursive models with linguistic, semantic, and reasoning priors acquired through large-scale pretraining. We further adapted \emph{HSW}, which assigns stronger supervision to earlier recursive steps while gradually reducing the contribution of later refinements. This weighting strategy improves optimization stability, reduces gradient variance, and encourages the model to make meaningful progress across the recursive trajectory. R-Qwen also remains robust when evaluated beyond its training horizon, yielding gains from additional test-time refinement. Across eight challenging symbolic reasoning benchmarks, R-Qwen consistently improves over strong baselines and demonstrates that pretrained autoregressive models can benefit substantially from explicit recursive refinement.

 \section*{Acknowledgements}

 Funded by the Natural Sciences and Engineering Research Council of Canada (NSERC), the Government of Canada’s New Frontiers in Research Fund (NFRF), [NFRFE-2022-00295], and Fonds Québécois de la Recherche sur la Nature et les Technologies group grant 361263.   OpenAI’s ChatGPT was used to assist with writing and editing. The authors reviewed and verified all content and accept full responsibility for the final manuscript. GL acknowledges support from the Canada-CIFAR AI Chair program.

\bibliography{aaai2027}


\clearpage

\section{Supplementary Material}
\label{SupplementaryMaterial}

\subsection{Extended Related Work}
\label{app:related_work}
\noindent\textbf{Recursive reasoning models.}
HRM uses two recurrent modules operating at different timescales, while TRM
simplifies this design to a single network that repeatedly refines a latent
prediction \citep{wang2025hrm,jolicoeurmartineau2025trm}. Later variants use
Mamba-2 operators or explicit symbol equivariance
\citep{wang2026tiny,freinschlag2026symbol}. The Recursive Stem Model (RSM)
retains a TRM-style backbone but detaches the recurrent-state history, treats
early iterations as warm-up computation, and applies supervision only at the
final step to learn a depth-agnostic transition operator
\citep{hakimi2026form}. Recursive Inference Machines (RIMs) instead provide a
broader inference-oriented formulation in which TRM is recovered as a special
case and can be extended with a reweighting mechanism inspired by classical
inference engines \citep{komisarczyk2026recursive}. Although effective on Sudoku,
ARC, and related structured benchmarks, these models are generally trained from
scratch for particular task families. Symbol-Equivariant Recurrent Reasoning
Models (SE-RRM)~\citep{freinschlag2026symbol} also require a specialized
position--symbol representation whose cost grows with the alphabet size.
R-Qwen instead uses a pretrained autoregressive backbone and represents each
recurrent state as an explicit serialized candidate.

The Fixed-Point Reasoning Model stabilizes looped Transformers and uses
fixed-point residuals for adaptive halting \citep{movahedi2026fixedpoint}. Its
reasoning state remains latent, however, and inference requires convergence
thresholds and damping. Several recent methods improve latent recursive models
through stochastic or multi-trajectory inference. Guided stochastic exploration
samples multiple latent trajectories and selects them using a stopping head
\citep{corbett2026boosting}, while Probabilistic TRM (PTRM) injects Gaussian
noise during recursion and uses TRM's existing Q head to select among parallel
rollouts \citep{sghaier2026probabilistic}. Generative Recursive reAsoning Models (GRAM)
learn stochastic latent transitions with amortized variational inference,
thereby representing multiple hypotheses and scaling computation through both
recursion depth and trajectory count \citep{baek2026generative}. The Energy-guided
Recursive Model (ERM) introduces an explicit Hopfield-energy criterion for
sampling, ranking, and selecting recursive trajectories
\citep{zhao2026energy}. These approaches provide stronger exploration than a
single deterministic latent rollout, but require multiple trajectories,
selection mechanisms, or probabilistic latent-state training. R-Qwen instead
performs deterministic, directly supervised candidate-to-candidate refinement
without a specialized recurrent architecture or trajectory-selection module.

An analysis of TRM on ARC-AGI-1 shows substantial dependence on large
test-time ensembles and learned puzzle identifiers, with most performance
obtained after the first recursion \citep{royeazar2025tiny}. R-Qwen does not use
puzzle-specific identity embeddings and exposes the effect of every refinement
step directly. Latent-State Supervised Reinforcement Learning instead decodes
intermediate latent states and applies process-supervised reinforcement learning
\citep{ren2025lsrl}; R-Qwen supervises complete candidates without an auxiliary
decoder.

\noindent\textbf{Looped and adaptive-depth Transformers.}
Universal Transformers and recurrent-depth models increase computation by
repeatedly applying shared blocks to hidden states
\citep{dehghani2019universal,geiping2025scaling}. Mixture-of-Recursions assigns
different recursive depths to individual tokens \citep{bae2025mixture}, but
requires architectural changes, learned routing, load balancing, and specialized
key--value caching. LoopFormer \citep{jeddi2026loopformer} and LOTUS
\citep{fan2026bridging} similarly operate over latent trajectories or latent
tokens and modify the network architecture. RecursiveVLM adapts parameter-shared
recursion specifically to large multimodal models through a recursive connector
that aligns intermediate vision and language features and a monotonic recursion
loss that supervises successive loops \citep{xu2026looping}. R-Qwen instead
performs recursion outside the backbone, allowing a pretrained model to refine
an interpretable answer and apply task constraints between steps.

Recent studies also examine how looped models acquire and control implicit
reasoning. Loop, Think, \& Generalize shows that recurrent-depth Transformers
can improve systematic generalization and extrapolate to reasoning depths beyond
those seen during training by increasing inference-time recurrence, while also
identifying degradation from excessive looping \citep{kohli2026loop}.
LoopRPT introduces reinforcement pre-training for looped language models by
assigning learning signals to intermediate latent steps rather than only to
output tokens \citep{tang2026looprpt}. These methods optimize computation within
persistent hidden states. R-Qwen instead serializes each intermediate candidate,
so every recursive step is readable, independently supervised, and re-encoded by
the pretrained backbone.

Dense supervision does not necessarily stabilize looped hidden states because
normalization can hide recurrent-state magnitude from the output loss
\citep{sharma2026dense}. R-Qwen avoids directly propagating a persistent hidden
state by serializing and re-encoding the candidate at every recursion.

\noindent\textbf{Continuous and fixed-point reasoning.}
Coconut \citep{hao2025training}, SeLaR~\cite{fu2026selar}, and
SpiralThinker~\cite{piao2026spiralthinker} perform reasoning partly or entirely
in continuous latent space. Attractor Models use an equilibrium solver, while
Flow Reasoning Models learn iterative discrete denoising dynamics with
verification and restarts \citep{feinashley2026solve,helbling2026flow}. These
approaches can support adaptive or parallel search but require specialized
architectures, latent-state training, or additional verification computation.
R-Qwen retains fully discrete candidates, enabling direct supervision and
deterministic enforcement of fixed clues, output alphabets, and grid structure.

\noindent\textbf{Autoregressive refinement and test-time adaptation.}
Tiny Autoregressive Recursive Models show that directly transferring TRM-style
latent recursion to next-token prediction is not consistently effective
\citep{rauba2026tiny}. R-Qwen instead formulates recursion as
prompt-conditioned candidate editing.

Test-time training adapts a model separately to each test task using its
demonstrations, augmentation, and task-specific LoRA optimization
\citep{akyurek2025surprising}. R-Qwen requires no gradient updates at inference
and reuses one learned refinement policy across tasks and instances.

Self-Refine and related self-correction methods rely on an LLM to generate
critiques before revising its answer
\citep{madaan2023selfrefine,he2025selfcorrection,wu2025visco}. Their success
depends strongly on critique quality. R-Qwen learns direct candidate correction
without a separate natural-language feedback stage. Recursive Language Models
recurse over decomposed input contexts \citep{zhang2026rlm}. Similarly,
Recursive Models for Long-Horizon Reasoning formalize self-invocation on isolated
subtasks as a way to overcome bounded active context and demonstrate this
strategy on long-horizon Boolean satisfiability \citep{yang2026recursive}.
These context-decomposition approaches differ from R-Qwen, which repeatedly
refines one evolving structured answer rather than constructing a recursive tree
of subproblems.

\noindent\textbf{Neuro-symbolic and visual reasoning.}
MaxSAT-based feedback uses an external solver to identify inconsistent Sudoku
assignments and guide a vision-language model \citep{orvalho2026maxsat}. It
provides stronger logical guarantees than R-Qwen's projection but requires a
task-specific symbolic encoding and repeated solver interaction. SATBench
similarly evaluates search over formally constrained assignments
\citep{wei2025satbench}.

Visual-reasoning methods use long reasoning trajectories, parameter merging,
modality switching, or visual test-time training
\citep{dong2025insightv,chen2025bring,zhang2026think,hu2026arc}. R-Qwen instead
uses one recursive candidate-refinement mechanism across visual, spatial,
symbolic, and language-grounded tasks without a specialized visual architecture,
external solver, or explicit long-chain rationale.


\subsection{Datasets}

We evaluate our method on four challenging symbolic reasoning benchmarks spanning constraint satisfaction, path planning, and abstract visual reasoning. These tasks require models to perform multi-step reasoning under strict logical constraints while generalizing beyond memorization.

\paragraph{Sudoku.}
We evaluate on the \emph{Sudoku-Extreme} benchmark introduced by \cite{palm2018recurrent}. Each instance consists of a partially filled $9\times9$ Sudoku grid, and the objective is to predict the unique valid completion satisfying the standard row, column, and $3\times3$ subgrid constraints. Compared with conventional Sudoku datasets, Sudoku-Extreme contains significantly more difficult puzzles that require long reasoning chains and extensive constraint propagation, making it a demanding benchmark for recursive reasoning models. During training, the model receives the partially observed grid as input and recursively refines its prediction until convergence.

\paragraph{Maze.}
For path planning, we use the Maze benchmark from \cite{lehnert2024beyond}. Each example consists of a randomly generated maze with designated start and goal locations, and the model must predict a valid path connecting them while avoiding obstacles. Unlike shortest-path algorithms that rely on explicit graph search, the benchmark evaluates whether a neural reasoning model can iteratively refine candidate solutions through recursive computation. Maze serves as a complementary benchmark to Sudoku by emphasizing sequential planning rather than constraint satisfaction.

\paragraph{ARC-1 and ARC-2.}
To evaluate abstract visual reasoning, we adopt the ARC-1 \cite{chollet2019arc} and ARC-2 benchmarks provided by \cite{arcprize2025arcagi2}. Each ARC task consists of several input-output demonstration pairs that define an underlying transformation rule, followed by a test input for which the correct output grid must be generated. The tasks require compositional reasoning, object manipulation, counting, geometric transformations, and abstraction over previously unseen concepts. ARC-AGI-2 contains substantially more challenging tasks than ARC-AGI-1 and was designed to reduce memorization effects, providing a stronger measure of systematic generalization.

\paragraph{CrossWordBench.}
We evaluate our method on CrossWordBench~\cite{leng2025crosswordbench}, a benchmark designed to assess reasoning through crossword puzzle solving in both text-only and vision-language settings. Each puzzle consists of natural-language clues together with a structured crossword grid, requiring the model to infer semantically correct answers while ensuring global consistency across intersecting words. The benchmark is generated using a controllable framework that supports multiple difficulty levels, enabling a comprehensive evaluation of reasoning ability. Following the original benchmark, we use the language-only version for our LLM experiments and the image-based version for our VLM experiments, allowing us to evaluate recursive reasoning in both textual and multimodal scenarios.

\paragraph{Vision Benchmarks} 
We evaluate our method on three vision-based abstract reasoning benchmarks: (1) the official 400-task ARC-AGI evaluation set~\cite{chollet2019arc}, (2) 100 randomly sampled tasks from Re-ARC~\cite{hodel2024addressing}, and (3) 100 randomly sampled tasks from BARC~\cite{li2025combining}. For both Re-ARC and BARC, each selected task consists of four input-output pairs, including three demonstration examples and one held-out test example used for evaluation.

\begin{table*}[t]
\caption{The five benchmarks as instances of serialized constraint
satisfaction. Only the serialization, clamped set, and context are
task-specific; the refinement operator, training objective, curriculum, and
HSW weighting are shared.}
\centering
\small
\begin{tabular}{lcccc}
\toprule
\textbf{Task} & \textbf{Alphabet $\Sigma$} & \textbf{Clamped set $\mathcal{G}(\bm{x})$} & \textbf{Context $c$} & \textbf{Target $\bm{y}^{\ast}$} \\
\midrule
Sudoku        & digits $1$--$9$        & given clue cells              & --                       & completed grid \\
Maze          & cell/path symbols      & walls, start, goal            & --                       & solution path overlay \\
ARC-AGI-1/2   & $10$ colours           & test input grid               & input/output demos       & output grid \\
CrossWordBench& letters $A$--$Z$       & prefilled letters             & across/down clues        & filled grid \\
\bottomrule
\end{tabular}

\label{tab:tasks}
\end{table*}

\begin{table}[ht]
\centering
\scriptsize
\caption{Training configurations on NVIDIA H100 GPUs.}
\label{tab:training_config}
\resizebox{.6\columnwidth}{!}{%
\begin{tabular}{@{}lccc@{}}
\toprule
\textbf{Task} & \textbf{Epochs} & \textbf{GPUs} & \textbf{Time} \\
\midrule
Sudoku & 10 & 4 & 22.5h \\
Maze & 10 & 4 & 3 days\\
ARC-1 & 40 & 4 & 4 days \\
ARC-2 & 40 & 4 & 2 days \\
ARC-AGI & 40 & 4 & 5 days \\
Re-ARC & 20 & 4 & 2 days \\
BARC & 20 & 4 & 2 days \\
\bottomrule
\end{tabular}}
\end{table}

\subsection{Training Configuration}

All experiments are implemented in PyTorch using the HuggingFace
\texttt{transformers} library together with the PEFT framework for
parameter-efficient fine-tuning.
We use a pretrained  Qwen3.5-27B language model as the backbone and
fine-tune only Low-Rank Adaptation (LoRA) parameters while keeping
all pretrained backbone weights frozen.
Unless otherwise stated, LoRA is applied to the query, key, and value
projection matrices of every self-attention layer with rank
$r=16$, scaling factor $\alpha=32$, and dropout $0.05$.
The optimizer is SGD with learning rate
$2\times10^{-4}$, weight decay $0.01$, cosine learning-rate decay,
and a linear warmup corresponding to $5\%$ of the total optimization
steps.
Gradient clipping with maximum norm $1.0$ is applied throughout
training.
Mixed-precision training using bfloat16 is employed whenever supported,
and the framework also supports optional 4-bit QLoRA training for
memory-efficient fine-tuning.
Training is performed on four NVIDIA H100 GPUs.
The batch size is task-dependent and chosen according to GPU memory.

\subsection{Recursive Training}

For every training sample, recursion is initialized using the original
problem itself as the first candidate,
$\hat{\mathbf{y}}^{(0)}=\mathbf{x}$.
At refinement step $t$, the model receives the original problem,
the current candidate solution,
and the current refinement index.
The model predicts an improved candidate, which is passed through the
task-specific projection function before becoming the input to the next
recursive step.
During supervised training, the model is optimized using teacher forcing,
where the target solution is generated while the prompt tokens are
masked from the loss.

Following our recursive training strategy, supervision is applied at
every refinement step rather than only at the final prediction.
The maximum recursion depth during training is
$N_{\max}=10$.
Instead of immediately training with the maximum depth,
we employ a supervision-depth curriculum that gradually increases the
number of supervised refinement steps according to the schedule

\[
(1\!:\!4,\;4\!:\!8,\;7\!:\!10),
\]

meaning that epochs $1$--$3$ use four recursive refinement steps,
epochs $4$--$6$ use eight steps,
and all subsequent epochs use ten steps.
This curriculum stabilizes optimization during the early stages of
training while exposing the model to longer recursive trajectories later
in training.

\begin{figure*}
  \centering
        \includegraphics[width=\textwidth]{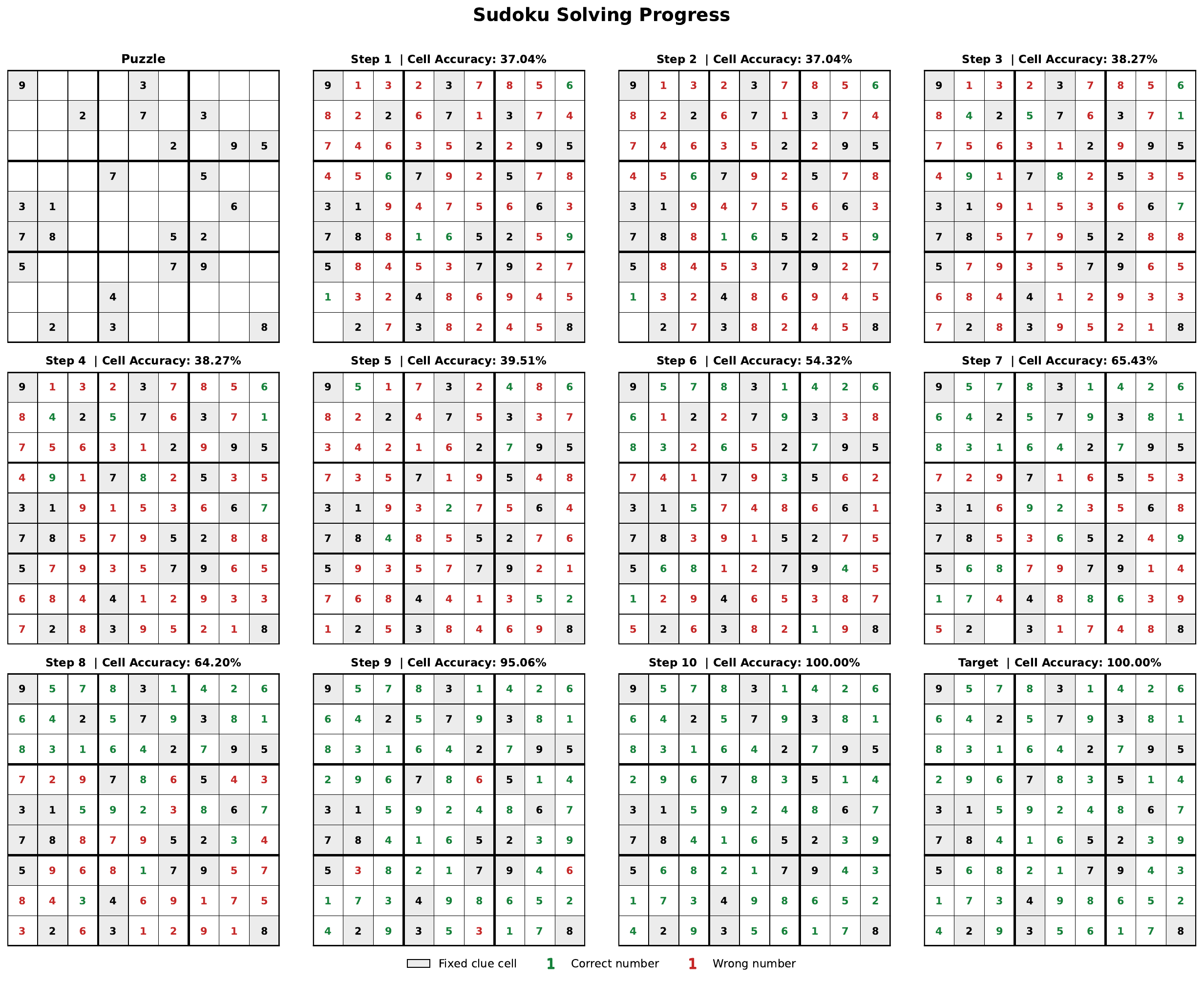}
        \caption{Visualization of a Sudoku puzzle and its solving steps using our R-Qwen model.}
    \label{fig:vis_sudoku}
\end{figure*}

\begin{figure*}
  \centering
        \includegraphics[width=\textwidth]{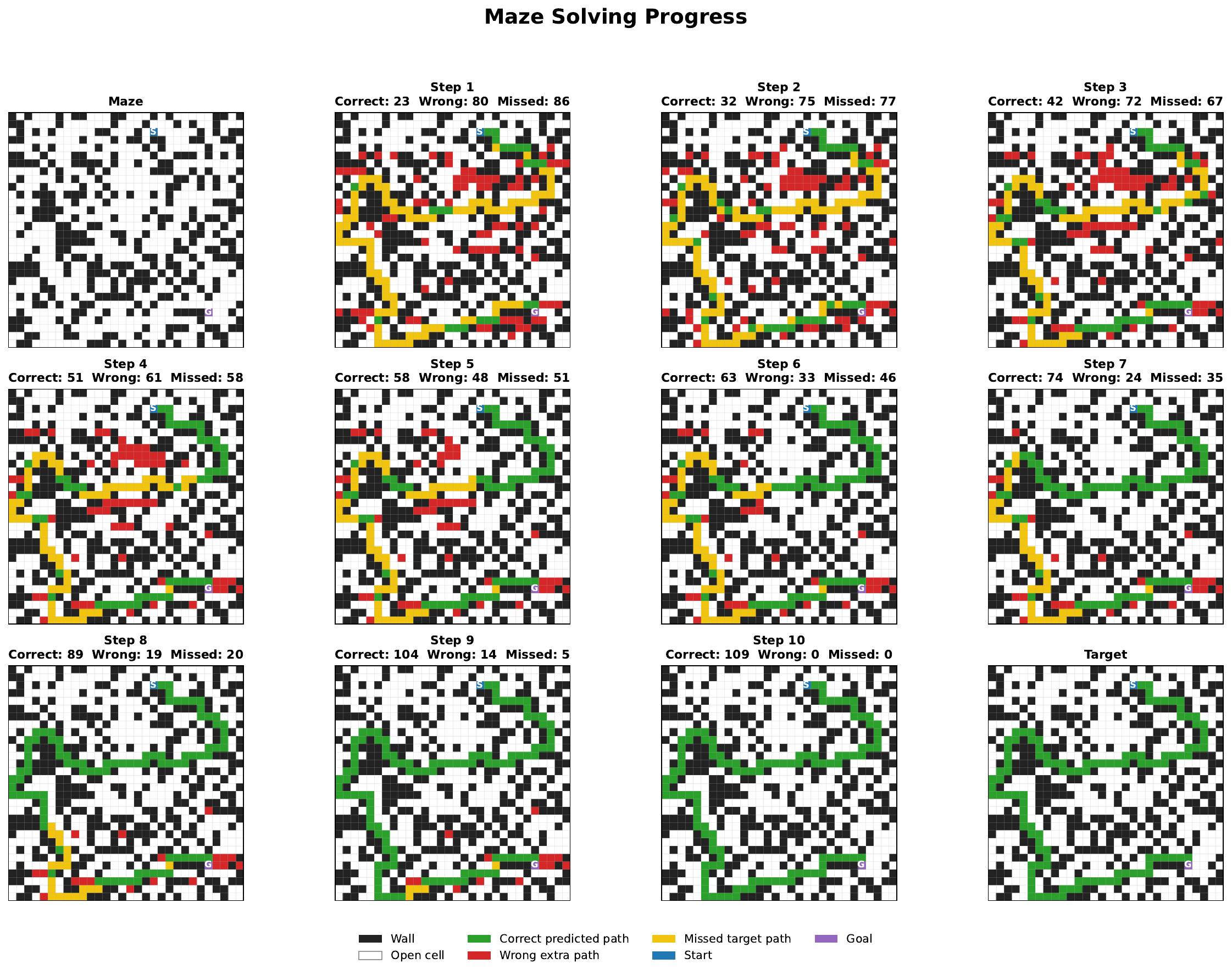}
        \caption{Visualization of a Maze puzzle and its solving steps using our R-Qwen model.}
    \label{fig:vis_maze}
\end{figure*}

\begin{figure*}
  \centering
        \includegraphics[width=\textwidth]{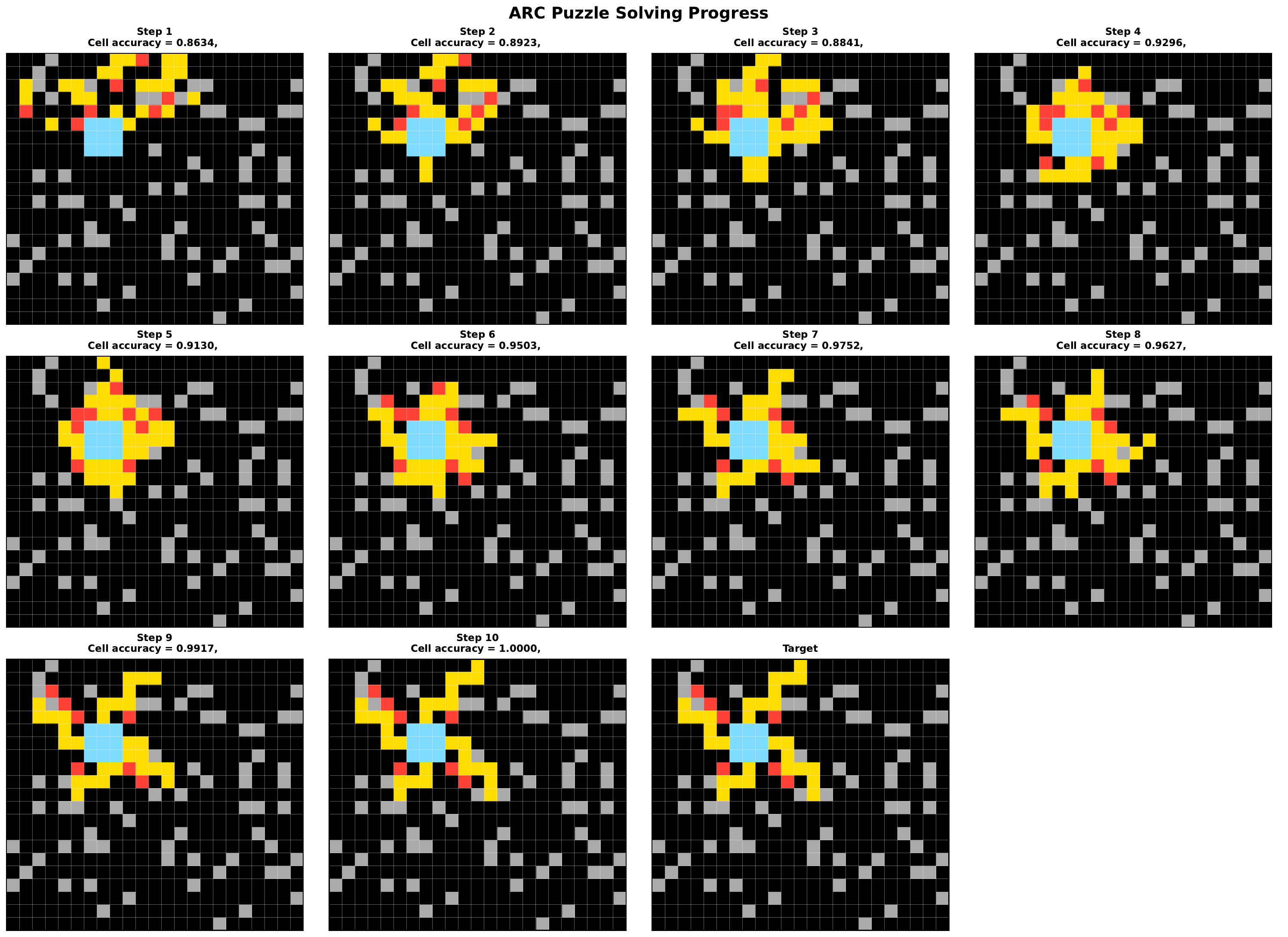}
        \caption{Visualization of a ARC puzzle and its solving steps using our R-Qwen model.}
    \label{fig:vis_arc}
\end{figure*}

\subsection{Visualization}
\paragraph{Sudoku solving process.}
Figure~\ref{fig:vis_sudoku} illustrates the step-by-step refinement of the Sudoku solution generated by R-Qwen. The original clues are shown in black, while correct and incorrect predictions are highlighted in green and red, respectively. During the first five iterations, the model makes limited progress, with cell accuracy increasing from \(37.04\%\) to \(39.51\%\). A more substantial improvement occurs in Steps~6 and~7, where the accuracy rises to \(54.32\%\) and \(65.43\%\), respectively. Although Step~8 exhibits a slight temporary decrease to \(64.20\%\), the model subsequently corrects most of the remaining errors, achieving \(95.06\%\) accuracy at Step~9 and \(100\%\) accuracy at Step~10. This non-monotonic refinement demonstrates that the model can revise previously generated values and progressively enforce the global constraints of the puzzle until its prediction exactly matches the target grid.

\paragraph{Maze solving process.}
Figure~\ref{fig:vis_maze} presents the recursive maze-solving process from the start position, denoted by \(S\), to the goal position, denoted by \(G\). Green cells indicate correctly predicted path segments, red cells represent unnecessary predicted segments, and yellow cells denote portions of the target path that have not yet been identified. In the first step, the predicted path is highly fragmented, containing only \(23\) correct cells, \(80\) incorrect extra cells, and \(86\) missed cells. Across successive refinement steps, the predicted route becomes increasingly aligned with the valid path: the number of correct cells steadily increases, while both the incorrect and missed path segments decrease. At Step~9, the model identifies \(104\) correct path cells, with only \(14\) extra cells and \(5\) missed cells. By Step~10, all \(109\) path cells are correctly recovered, with no extra or missing segments, producing a solution identical to the target path.

\paragraph{ARC puzzle solving process.}
Figure~\ref{fig:vis_arc} visualizes the iterative solution of an ARC task, in which the model must infer the underlying transformation rule and reconstruct the target spatial pattern. Starting from a partially correct configuration with a cell accuracy of \(86.34\%\), the model progressively reorganizes the relevant colored cells while preserving the surrounding background and distractor elements. The refinement process is not strictly monotonic, as small decreases in accuracy occur at Steps~3, 5, and~8. Nevertheless, the overall structure becomes increasingly consistent with the target, reaching \(95.03\%\) accuracy at Step~6, \(97.52\%\) at Step~7, and \(99.17\%\) at Step~9. In the final refinement step, the model corrects the remaining erroneous cells and achieves \(100\%\) cell accuracy, exactly reproducing the target pattern. This progression highlights the model's ability to revise intermediate hypotheses and gradually converge toward the correct abstract transformation.

\subsection{Pseudocode}
\begin{algorithm}[H]
\caption{HSW-Guided Recursive LoRA Fine-Tuning}
\label{alg:hsw_recursive_qwen}
\begin{algorithmic}[1]

\Require
Multi-task training set
$\mathcal{D}=\{(d_i,\bm{x}_i,c_i,\bm{y}^{*}_i)\}$;
pretrained Qwen model $\pi_{\theta_0}$;
LoRA parameters $\Delta\theta$;
maximum supervision depth $N_{\max}$;
curriculum schedule
$\{(e_k,N_k)\}_{k=1}^{S}$;
HSW decay factor $\lambda$;
generation interval $m$

\Ensure
Fine-tuned LoRA parameters $\Delta\theta$

\State Freeze pretrained parameters $\theta_0$
\State Insert trainable LoRA adapters $\Delta\theta$
\State Initialize optimizer and learning-rate scheduler

\For{epoch $e=1$ to $E$}

    \State Determine active recursion horizon:
    \[
    N \gets N_{\mathrm{sup}}(e)
    \]

    \State Compute HSW normalization:
    \[
    Z_{\lambda}
    \gets
    \sum_{s=1}^{N}\lambda^{s-1}
    \]

    \For{each mini-batch
    $\mathcal{B}=\{(d_i,\bm{x}_i,c_i,\bm{y}^{*}_i)\}_{i=1}^{B}$}

        \For{each instance $i$}
            \State Initialize candidate:
            \[
            \hat{\bm{y}}_i^{(0)}\gets\bm{x}_i
            \]
        \EndFor

        \State Initialize active set:
        \[
        \mathcal{A}_1
        \gets
        \{i:\hat{\bm{y}}_i^{(0)}
        \neq\bm{y}^{*}_i\}
        \]

        \For{$t=1$ to $N$}

            \If{$\mathcal{A}_t=\varnothing$}
                \State \textbf{break}
            \EndIf

            \State Compute the HSW step weight:
            \[
            w_t
            \gets
            \frac{\lambda^{t-1}}{Z_{\lambda}}
            \]

            \For{each active instance $i\in\mathcal{A}_t$}

                \State Construct the candidate-conditioned prompt:
                \[
                p_i^{(t)}
                \gets
                P(
                \bm{x}_i,
                c_i,
                \hat{\bm{y}}_i^{(t-1)},
                t,
                N)
                \]

                \State Tokenize
                $p_i^{(t)}\Vert\bm{y}^{*}_i$

                \State Mask prompt and padding tokens in the labels

            \EndFor

            \State Compute target-only cross-entropy:
            \[
            \ell^{(t)}
            =
            -\frac{1}{|\mathcal{A}_t|}
            \sum_{i\in\mathcal{A}_t}
            \frac{1}{|\bm{y}^{*}_i|}
            \sum_{j=1}^{|\bm{y}^{*}_i|}
            \log
            \pi_{\theta_0,\Delta\theta}
            \left(
            y^{*}_{i,j}
            \mid
            p_i^{(t)},
            \bm{y}^{*}_{i,<j}
            \right)
            \]

            \State Apply Hierarchical Supervision Weighting:
            \[
            \mathcal{L}^{(t)}
            \gets
            w_t\,\ell^{(t)}
            \]

            \State Backpropagate $\mathcal{L}^{(t)}$
            \State Clip gradients
            \State Update $\Delta\theta$ using the optimizer
            \State Update the learning-rate scheduler
            \State Reset gradients

            \If{$t\bmod m=0$ \textbf{or} $t=N$}

                \For{each active instance $i\in\mathcal{A}_t$}

                    \State Generate an improved candidate:
                    \[
                    \tilde{\bm{y}}_i^{(t)}
                    \gets
                    \operatorname{GreedyDecode}
                    \left(
                    \pi_{\theta_0,\Delta\theta},
                    p_i^{(t)}
                    \right)
                    \]

                    \State Apply the task-specific projection:
                    \[
                    \hat{\bm{y}}_i^{(t)}
                    \gets
                    \Pi_{\bm{x}_i}
                    \left(
                    \tilde{\bm{y}}_i^{(t)};
                    \hat{\bm{y}}_i^{(t-1)}
                    \right)
                    \]

                    \State Detach $\hat{\bm{y}}_i^{(t)}$
                    from the computation graph

                \EndFor

            \Else

                \For{each active instance $i\in\mathcal{A}_t$}
                    \State Carry the candidate forward:
                    \[
                    \hat{\bm{y}}_i^{(t)}
                    \gets
                    \hat{\bm{y}}_i^{(t-1)}
                    \]
                \EndFor

            \EndIf

            \State Update the active set:
            \[
            \mathcal{A}_{t+1}
            \gets
            \left\{
            i\in\mathcal{A}_t:
            \hat{\bm{y}}_i^{(t)}
            \neq
            \bm{y}^{*}_i
            \right\}
            \]

        \EndFor
    \EndFor

    \State Evaluate using fixed-step recursive inference
    \State Save the LoRA adapters and validation metrics

\EndFor

\State \Return $\Delta\theta$

\end{algorithmic}
\end{algorithm}

\begin{table*}[t]
\centering
\caption{
Comparison with recent inference-time reasoning approaches.
We report terminal exact-solve accuracy (\%). Results are separated
by dataset because Sudoku (Shah) and Sudoku-Extreme represent
different evaluation distributions. Test-time computation also
differs substantially across methods.
}
\label{tab:recent_inference_comparison}
\resizebox{\textwidth}{!}{
\begin{tabular}{lccccc}
\toprule
\textbf{Method}
& \textbf{\# Params}
& \textbf{Test-time configuration}
& \textbf{Sudoku (Shah)}
& \textbf{Sudoku-Extreme}
& \textbf{Maze-Hard} \\
\midrule

Flow Reasoning Model (FRM)
\cite{helbling2026flow}
& $\sim$30M
& Verifier-gated restart, $\sim$7 NFE
& 99.2
& --
& -- \\

FRM + self-verification scaling
\cite{helbling2026flow}
& $\sim$30M
& 8,192 rounds
& --
& 96.1
& -- \\

Base Flow + self-verification scaling
\cite{helbling2026flow}
& $\sim$30M
& 8,192 rounds
& --
& 98.6
& -- \\

Guided MAP
\cite{corbett2026boosting}
& N/R
& 16 particles, 48 outer steps
& --
& $98.0{\pm}0.3$
& $85.3{\pm}2.5$ \\

\rowcolor{green!3}
Ours
& 27B (12M$^\dagger$)
& Recursive refinement
& --
& 88.5
& \textbf{93.1} \\

\bottomrule
\end{tabular}
}
\vspace{1mm}

\footnotesize
$^\dagger$Number of trainable parameters.
NFE denotes the number of model forward evaluations.
N/R indicates that the parameter count of the specific frozen
Nano-TRM checkpoint was not restated in the corresponding paper.
Each self-verification round in FRM uses a 128-step proposal;
therefore, the 8,192-round results involve substantially greater
test-time computation than a standard single prediction.
\end{table*}

\paragraph{Comparison with recent inference-time scaling methods.}
Table~\ref{tab:recent_inference_comparison} compares our method
with two concurrent approaches that allocate additional computation
during inference. Helbling et al.~\cite{helbling2026flow} report
99.2\% accuracy on the in-distribution Sudoku dataset of Shah et al.
using approximately seven model evaluations. This result is not
directly comparable with Sudoku-Extreme. On the out-of-distribution
Sudoku-Extreme benchmark, their FRM with self-verification reaches
96.1\% using 8,192 verification-and-restart rounds, while the
corresponding base-flow configuration reaches 98.6\%.

Corbett et al.~\cite{corbett2026boosting} apply guided stochastic
exploration to a frozen TRM model without retraining its backbone.
Using 16 particles and 48 outer recursion steps, Guided MAP improves
Sudoku-Extreme accuracy from 85.9\% to 98.0\%. However, the learned
guide does not provide a corresponding improvement on Maze-Hard,
where Guided MAP obtains 85.3\%. In comparison, our method achieves
93.1\% on Maze-Hard, while jointly supporting Sudoku, Maze, ARC-1,
and ARC-2 within a unified recursive language-model framework.
These comparisons should be interpreted with consideration of the
different model architectures, training distributions, and
test-time computation budgets.

\end{document}